\documentclass[letterpaper, 9 pt, conference]{ieeetran}
\IEEEoverridecommandlockouts
\pdfoutput=1

\usepackage{cite}
\usepackage{amsmath,amssymb,amsfonts}
\usepackage{algorithmic}
\usepackage{graphicx}
\usepackage{textcomp}
\usepackage{xcolor}
\usepackage{svg}
\usepackage{multirow}
\usepackage{algorithm}
\usepackage{bm}
\usepackage{booktabs}
\usepackage{todonotes}
\usepackage[font=scriptsize]{caption} 

\usepackage{lipsum}
\usepackage{float}

\title{\LARGE \bf
Fully Spiking Neural Network for Legged Robots
}

\author{\IEEEauthorblockN{Xiaoyang Jiang$^{1,}$\IEEEauthorrefmark{1},
Qiang Zhang$^{2,}$\IEEEauthorrefmark{1},
Jingkai Sun$^{2}$, 
Jiahang Cao$^{2}$, 
Jingtong Ma$^{3}$,
Renjing Xu$^{2,}$\IEEEauthorrefmark{2}}

\IEEEauthorblockA{$^{1}$Center of Data Science, New York University, New York City, USA}
\IEEEauthorblockA{$^{3}$Center of Biomedical Engineering, Duke university, Durham, USA}
\IEEEauthorblockA{$^{2}$Function Hub, The Hong Kong University of Science and Technology (Guangzhou), Guangzhou, China}
\thanks{$^{*}$equal contributors, $^{\dagger}$corresponding author (renjingxu@hkust-gz.edu.cn)}}

\def\BibTeX{{\rm B\kern-.05em{\sc i\kern-.025em b}\kern-.08em
    T\kern-.1667em\lower.7ex\hbox{E}\kern-.125emX}}

\begin{document}



\maketitle
\thispagestyle{empty}
\pagestyle{empty}

\vspace{-0.37cm}
\begin{abstract}
Recent advancements in legged robots using deep reinforcement learning have led to significant progress. Quadruped robots can perform complex tasks in challenging environments, while bipedal and humanoid robots have also achieved breakthroughs. Current reinforcement learning methods leverage diverse robot bodies and historical information to perform actions, but previous research has not emphasized the speed and energy consumption of network inference and the biological significance of neural networks. Most networks are traditional artificial neural networks that utilize multilayer perceptrons (MLP). This paper presents a novel Spiking Neural Network (SNN) for legged robots, showing exceptional performance in various simulated terrains. SNNs provide natural advantages in inference speed and energy consumption, and their pulse-form processing enhances biological interpretability. This study presents a highly efficient SNN for legged robots that can be seamless integrated into other learning models.
\end{abstract}

\section{Introduction}

The increasing adoption of mobile robots with continuous high-dimensional observations and action spaces necessitates advanced control algorithms for complex real-world tasks. Currently, the limited onboard energy resources hinder continuous and cost-effective operation, creating an urgent need for energy-efficient solutions for the seamless control of these robots. Deep reinforcement learning (DRL) employs deep neural networks (DNNs) as potent function approximators for learning optimal control strategies for intricate tasks \cite{ha2018automated,zhu2017target}, through mapping the original state space to the action space \cite{duan2016benchmarking,lillicrap2015continuous}. \cite{ho2016generative} differs from traditional reinforcement learning by imitating behaviors from reference datasets through a generative adversarial network. Adversarial Motion Priors (AMP) \cite{peng2021amp} enhances \cite{ho2016generative} by combining task and imitation rewards, enabling agents to mimic actions from reference datasets. To learn from unlabeled references dataset, \cite{li2023versatile} employs a skill discriminator, allowing quadrupeds to master various gaits and perform backflips. Furthermore, \cite{wu2023learning} integrates Rapid Motor Adaptation (RMA) with AMP, improving quadrupeds' ability to traverse challenging terrains rapidly. While DRL delivers impressive performance, it often incurs high energy consumption and slower execution speeds. DNN-based control strategies generally operate slower than motion units, causing step-like control signals that hinder performance. 

\begin{figure}[tbp]
\vspace{0.1cm}
\setlength{\belowcaptionskip}{-0.6cm}
\centerline{\includegraphics[width=\columnwidth]{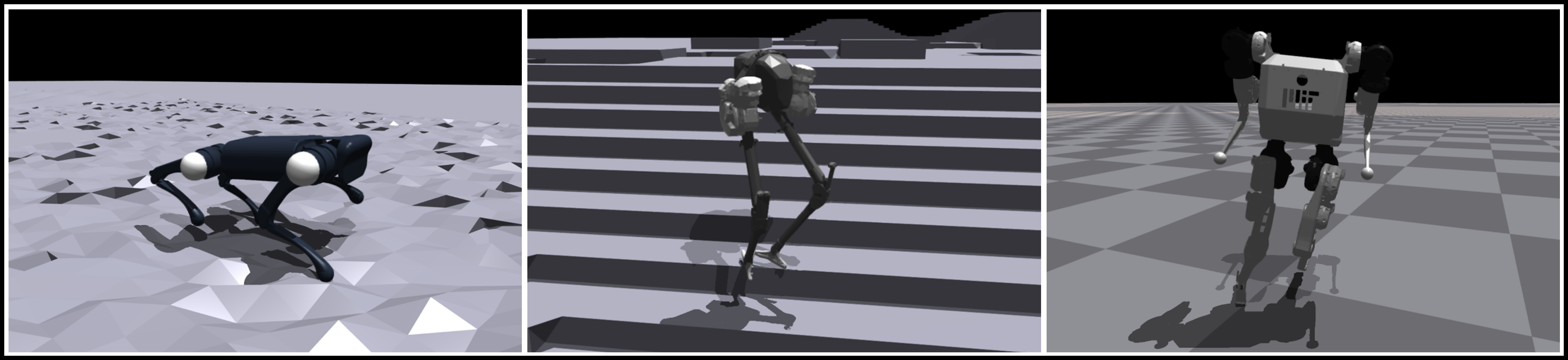}}
\caption{Whole-body control on various types of robots through our spike-based approach. This innovative methodology allows us to effectively regulate and coordinate the robots' movements, enhancing their overall performance and versatility. \textbf{Left}: A1 \textbf{Middle}: Cassie \textbf{Right}: MIT Humanoid}
\label{fig1}
\end{figure}

Spiking neural networks, or third-generation neural networks, provide an energy-efficient and high-speed alternative for deep learning by utilizing neuromorphic computing principles \cite{davies2018loihi}. The biological plausibility, the significant increase in energy efficiency (particularly when deployed on neuromorphic chips \cite{roy2019towards}), high-speed processing and real-time capability for high-dimensional data (especially from asynchronous sensors like event-based cameras \cite{gallego2020event}) contribute to the advantages that SNNs possess over ANNs in specific applications. Recently, many works have grown up around introducing SNNs into RL algorithms\cite{florian2007reinforcement,o2013spiking,yuan2019reinforcement,doya2000reinforcement,fremaux2013reinforcement}. Research shows that SNNs are energy-efficient and high-speed solutions for robot control in scenarios with limited onboard energy resources \cite{tang2019spiking,taunyazov2020event,michaelis2020robust}. To address the limitations of SNNs in high-dimensional control problems, combining their energy efficiency with the optimality of DRL offers a promising solution, as DRL has proven effective in various control tasks \cite{mnih2015human}. 
Rewards act as training guides in DRL, some studies utilize a three-factor learning rule \cite{fremaux2013reinforcement}. While effective in low-dimensional tasks, these rules struggle with complex problems, complicating optimization without a global loss function \cite{legenstein2005can}. Recently, \cite{rosenfeld2019learning} proposed a strategy gradient-based algorithm for training a SNN to learn random strategies, but it is limited to discrete action spaces, hindering its use in high-dimensional continuous control problems.

The recent conceptualization of the brain's topology and computational principles has ignited advancements in SNNs, exhibiting both human-like behavior\cite{balachandar2020spiking} and superior performance\cite{kreiser2020chip}. A key feature of brain computation is the use of neuronal populations to encode information, from sensory input to output signals. Each neuron has a receptive field that captures a specific segment of the signal \cite{georgopoulos1986neuronal}. Notably, initial investigations into this group coding scheme have shown its enhanced capability to represent stimuli\cite{tkavcik2010optimal}, contributing to triumphs in training SNNs for complex, high-dimensional supervised learning tasks\cite{bellec2018long,pan2019neural}. 
The main contributions of this paper can be summarized as follows:
\begin{itemize}
    \item For the first time, we have implemented a lightweight population coded SNNs on a policy network in various legged robots simulated in Isaac Gym\cite{makoviychuk2021isaac} using a multi-stage training method. 
    We also integrated this method with imitation learning and trajectory history, achieving effective training outcomes.
    \item Our approach presents a considerable advantage over ANNs in terms of energy efficiency. This advantage holds substantial significance for enhancing the structural integrity and reducing the costs associated with robot development. 
    \item The research affirms the exceptional performance demonstrated by SNNs in high-frequency robot control, coupled with their significant edge over ANNs in attenuating signal noise, which enhances their robustness in practical situations.
\end{itemize}

\begin{figure}[tbp]
\setlength{\abovecaptionskip}{0cm}
\setlength{\belowcaptionskip}{-0.4cm}
\centerline{\includegraphics[width=1.02\columnwidth]{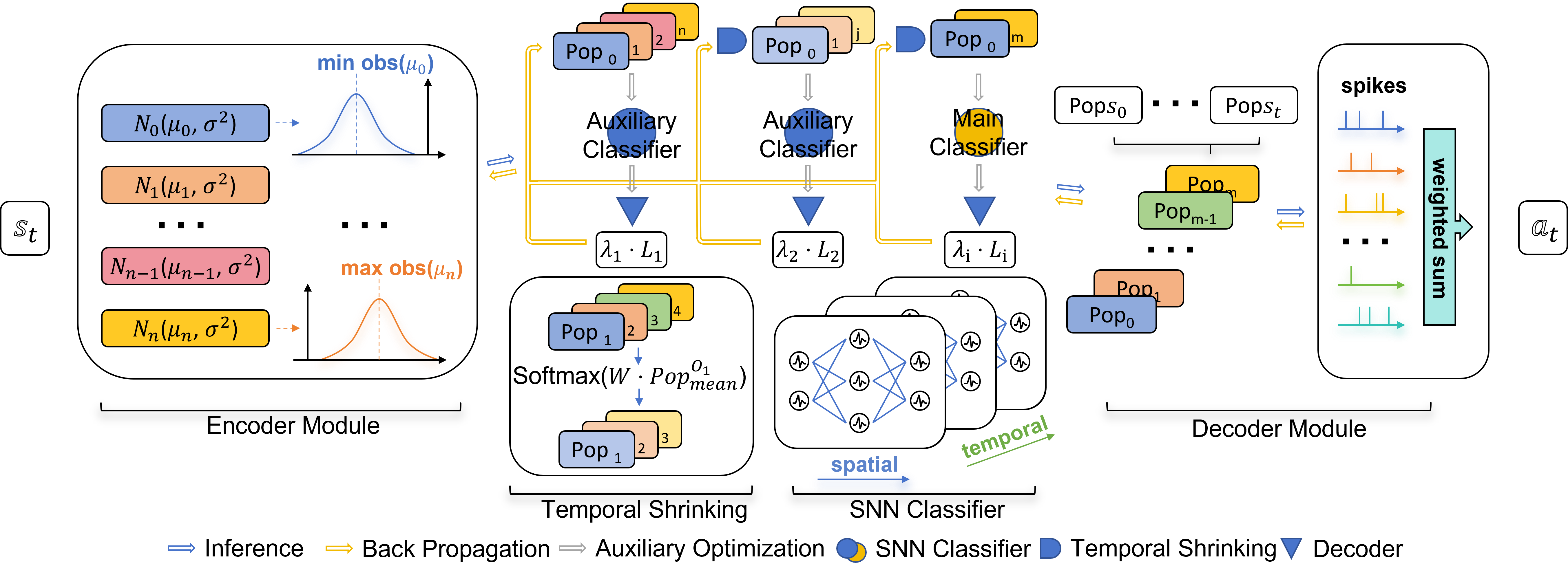}}
\caption{The observations are initially encoded by the encoder as $n$ independent distributions that are uniformly distributed over the observation range. After encoding, the population processes the distributions, resulting in spike generation. The neurons in the input populations encode each observation dimension and drive a multi-layered, fully connected SNN. During forward timesteps in PopSAN, the activities of each output population are decoded to determine the corresponding action dimension. The neural network receives observations, processes them using the SNN, and decodes the resulting activities to determine the appropriate action for the specific situation.}
\label{fig2}
\end{figure}

\section{Methods}


\subsection{SNN based Policy Network}




We employ a population-coded spiking actor-network (PopSAN)\cite{tang2021deep} that is trained in tandem with a deep critic network using the DRL algorithms. During training, the PopSAN generated an action $\alpha$ $\in$ $\mathbb{R}^{N}$ for a given observation, $s$, and the deep critic network predicted the associated state value $V$($s$) or action-value $Q$($s$, $\alpha$), which in turn optimized the PopSAN, in accordance with a chosen DRL method (Fig. \ref{fig2}). Within the PopSAN architecture, the encoder module encodes individual dimensions of the observation by mapping them to the activity of distinct neuron populations. During forward propagation, the input populations activate a multi-layer fully-connected SNN, producing activity patterns within the output populations. After each set of $T$ timesteps, these patterns of activity are decoded to ascertain the associated action dimensions.

The current-based leaky-integrate-and-fire (LIF) model of a spiking neuron is employed in constructing the SNN. The dynamics of the LIF neurons are governed by a two-step model:
i) the integration of presynaptic spikes $o$ into current $c$; and ii) the integration of current $c$ into membrane voltage $v$; $d_c$ and $d_v$ represent the current and voltage decay factors, respectively. In this implementation, a neuron fires a spike when its membrane potential surpasses a predetermined threshold. The hard-reset model was implemented, in which the membrane potential is promptly reset to the resting potential following a spike. Resultant spikes are transmitted to post-synaptic neurons during the same inference timestep, assuming zero propagation delay.
This approach facilitates efficient and synchronized information transmission within the SNN.


\subsection{Temporal Shrinking}

Next, inspired by \cite{ding2024shrinking}, we process the encoded information from the encoder in i stages. At each subsequent stage, the time step can be reduced by an arbitrary scale. Assuming that the time step for the first stage is $T_1(T_1 = n)$ and the scale for the second stage is $T_2(T_2 = j)$, we utilize a learnable weight $W$ $\in$ $\mathbb{R}^{T_2 \times T_1}$ to perform the scale conversion, as illustrated below:
\begin{equation}
\setlength\abovedisplayskip{4pt}
\setlength\belowdisplayskip{4pt}
\begin{aligned}
I_2 = O_1 \odot Softmax(WPop_{mean}^{O_1})
\label{eq1}
\end{aligned}
\end{equation}
Where $I_1$ $\in$ $\mathbb{R}^{T_1 \times obs}$ represents the input of the first stage, which is at the same scale as the output $O_1$ $\in$ $\mathbb{R}^{T_1 \times obs}$ of the first stage, while the input of the second stage should be $I_2$ $\in$ $\mathbb{R}^{T_2 \times obs}$, and $Pop_{mean}^{O_1}$ $\in$ $\mathbb{R}^{T_1 \times 1}$ means the average of $O_1$ at each time point:
\begin{equation}
\setlength\abovedisplayskip{4pt}
\setlength\belowdisplayskip{4pt}
\begin{aligned}
Pop_{mean}^{O_1} = \frac{1}{obs}\sum_{p=0}^{obs}O_{1,p}
\label{eq2}
\end{aligned}
\end{equation}
The $obs$ in \eqref{eq2} refers to the number of elements observed at a single time step. Equation \eqref{eq1} utilize softmax function to ensure that the probabilities allocated at all time steps sum to 1, ensuring information integrity. This way, \eqref{eq2} can be seen as a guidance for \eqref{eq1} to learn how to allocate information from stage 1 to stage 2. The method enables infinite time step compression to a lightweight level at minimal cost, greatly affecting low-latency needs in robot gait control.

\subsection{Auxiliary Optimization}

The use of surrogate gradient in training SNNs effectively tackles the non-differentiability of spikes, yet the discrepancy with the true gradient poses a limitation on SNN performance. Moreover, spikes face serious gradient vanishing/exploding problems \cite{fang2021deep}. To alleviate the aforementioned issues and ensure the effectiveness of temporal shrinking, we introduce auxiliary optimization. After each stage (except the final stage), $O_i$ is fed into the auxiliary classifier to obtain the stage loss in addition to the subsequent stage. The auxiliary classifier consists of an SNN classifier and a decoder module, where the time dimension of the SNN is equal to $T_i$ of the corresponding stage. The overall loss $L_{all}$ should be represented as a weighted sum of the losses from each stage, expressed as follows:
\begin{equation}
\setlength\abovedisplayskip{4pt}
\setlength\belowdisplayskip{4pt}
\begin{aligned}
L_{all} = \sum_i^{I}\lambda_iL_i(Y_i, \mathbb{S}_i), \sum_i^{I}\lambda_i = 1
\label{eq3}
\end{aligned}
\end{equation}
Where $\lambda_i$ is a manually set parameter representing the weight of each stage. $Y_i$ denotes the output of the auxiliary classifier in stage $i$, while $L_i$ represents the loss gained by interacting $Y_i$ with the environment $\mathbb{S}_i$ at this stage. By adjusting $\lambda_i$, we can emphasize either the global action features or the details of specific actions.

\subsection{Combination of Other Methods}

\begin{figure}[tbp]
\setlength{\belowcaptionskip}{-0.4cm}
\centerline{\includegraphics[scale=0.102]{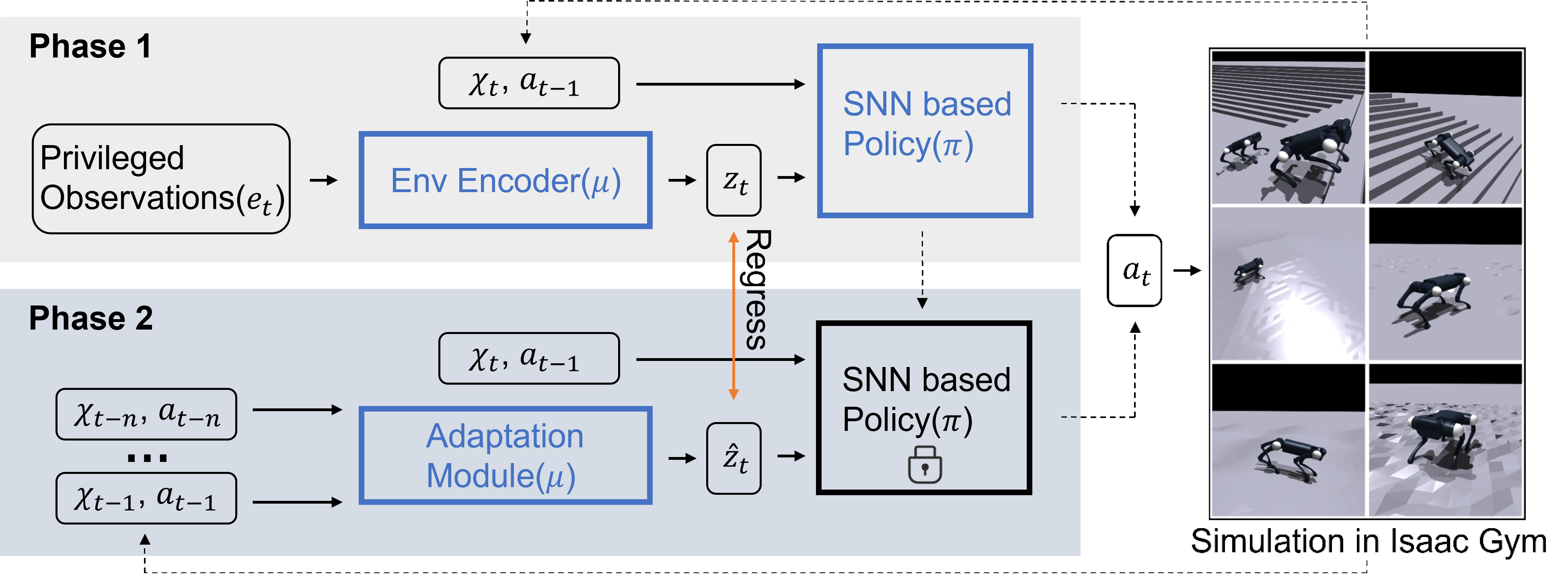}}
\caption{RMA consists of two subsystems: the base policy $\pi$ and the adaptation module $\phi$. The RMA training consists of two phases. \textbf{Training the Base Policy (Phase 1):} In the initial phase, the base policy $\pi$ is trained using PopSAN. The system takes the current state $x_t$, the previous action $\alpha_{t-1}$, and the environmental factors $e_t$ as input. These environmental factors are encoded into a latent extrinsics vector $z_t$ using the environmental factor encoder $\mu$. \textbf{Training the Adaptation Module (Phase 2):}In the second phase, the adaptation module $\phi$ is trained to predict the extrinsics $\widehat{z_t}$ using past states and actions. This training utilizes supervised learning with on-policy data. The adaptation module learns to capture the relationship between the state-action history and the corresponding extrinsics. 
}
\label{fig3}
\end{figure}

To validate the generalizability of our method, we further combined the advantages of SNN with the advanced RMA algorithm. Figure \ref{fig3} shows that the RMA system consists of two interconnected subsystems: the base policy $\pi$ and the adaptation module $\phi$. 
The base policy is trained using reinforcement learning in simulation, leveraging privileged information about the environment configuration $e_t$, such as friction, payload, and other factors. By utilizing the vector $e_t$, the base policy can adapt effectively to the unique characteristics of the environment. 
The purpose of $\phi$ is to estimate the extrinsics vector $z_t$ based solely on the recent state and action history of the robot, without direct access to $e_t$. 

\begin{figure}[tbp]
\setlength{\abovecaptionskip}{0.05cm}
\setlength{\belowcaptionskip}{-0.43cm}
\centerline{\includegraphics[width=0.96\columnwidth]{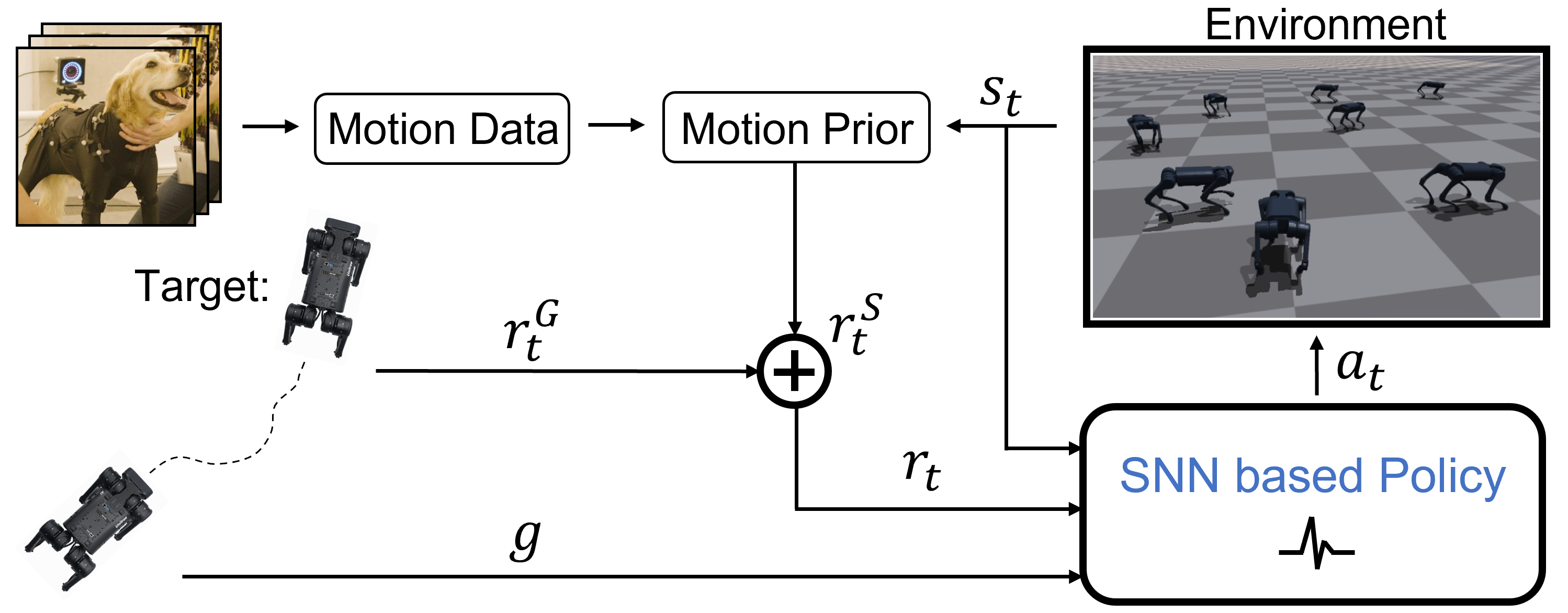}}
\caption{By leveraging Adversarial Motion Priors and employing PopSAN as a replacement for the policy network during training, the agent is able to generate behaviors that capture the essence of the motion capture dataset.}
\label{fig4}
\end{figure}

In addition, we have successfully combined SNN with AMP and achieved similar performance to ANN on legged robots. Figure \ref{fig4} provides a schematic overview of the system. The motion dataset $M$ consists of a collection of reference motions, where each motion $m^i$ $=$ {$\widehat{q}_{t}^{i}$} is represented as a sequence of poses $\widehat{q}_{t}^{i}$. 
The simulated robot's movement is governed by a policy $\pi(\alpha_t | s_t,g)$ that links the character's state $s_t$ and a given goal $g$ to a distribution of actions $\alpha_t$. The policy generates actions that determine the desired target positions for proportional-derivative (PD) controllers at each joint of the robot. The controllers then generate control forces to propel the robot's motion according to specified target positions. The goal $g$ defines a task reward function $r_{t}^{G} = r^{G}(s_t, \alpha_t, s_{t+1}, g)$ that outlines the high-level objectives the robot needs to achieve. The style objective $r_{t}^{S} = r^{S}(s_t, s_{t+1})$ is determined by an adversarial discriminator, which provides an a priori estimate of the naturalness or style of a motion, independent of the task at hand. By doing this, the style objective encourages the policy to produce movements that closely mirror the behaviors seen in the dataset.

\subsection{Training}

In our study, we used gradient descent to update the PopSAN parameters, with the specific loss function depending on the algorithm chosen (RMA or AMP). To train PopSAN parameters, we use the gradient of the loss with respect to the computed action, denoted as $\nabla_a$$L$. The parameters for each output population $i, i \in 1, ..., M$ are updated independently as follows:
\begin{equation}
\setlength\abovedisplayskip{4pt}
\setlength\belowdisplayskip{4pt}
\begin{aligned}
\nabla_{{\bm{W}_d}^{(i)}}L = \nabla_{\alpha_i}L\cdot{\bm{W}_d}^{(i)}\cdot\bm{fr}^{(i)}, \nabla_{{b_d}^{(i)}}L = \nabla_{\alpha_i}L\cdot{\bm{W}_d}^{(i)}
\label{eq4}
\end{aligned}
\end{equation}
The SNN parameters are updated using extended spatiotemporal backpropagation as introduced in \cite{tang2020reinforcement}. We utilized the rectangular function $z(v)$, as defined in \cite{wu2018spatio}, to estimate the spike's gradient. The gradients of the loss with respect to the parameters of the SNN for each layer $k$ are computed by aggregating the gradients backpropagated from all timesteps:
\begin{equation}
\setlength\abovedisplayskip{3pt}
\setlength\belowdisplayskip{4pt}
\begin{aligned}
\nabla_{{\bm{W}}^{(k)}}L=\sum_{t=1}^T\bm{o}^{(t)(k-1)}\cdot\nabla_{{\bm{c}}^{(t)(k)}}L, \nabla_{{\bm{b}}^{(k)}}L=\sum_{t=1}^T\nabla_{{\bm{c}}^{(t)(k)}}L 
\label{eq5}
\end{aligned}
\end{equation}
Lastly, we updated the parameters independently for each input population $i, i \in 1, ..., N$ as follows: 
\begin{equation}
\setlength\abovedisplayskip{4pt}
\setlength\belowdisplayskip{4pt}
\begin{aligned}
\nabla_{{\bm{\mu}}^{(i)}}L=&\sum_{t=1}^T\nabla_{{\bm{o}_i}^{(t)(o)}}L\cdot\bm{A_E}^{(i)}\cdot\frac{s_i - \bm{\mu}^{(i)}}{\bm{\sigma^{(i)^2}}}, \\
\nabla_{{\bm{\sigma}}^{(i)}}L=&\sum_{t=1}^T\nabla_{{\bm{o}_i}^{(t)(o)}}L\cdot\bm{A_E}^{(i)}\cdot\frac{{(s_i - \bm{\mu}^{(i)})}^2}{\bm{\sigma^{(i)^3}}}
\label{eq6}
\end{aligned}
\end{equation}

\section{Experiments}

The objectives of our experiments are the followings: \textrm{i}) To validate the feasibility of SNNs on robots operating in environments with high dimensions and intricate dynamics models. \textrm{ii}) Verify the benefits of SNNs over ANNs in terms of ultra-high frequency control. We assessed our approach using the Isaac Gym platform, which is a simulation platform specifically created for robotics applications.
We primarily evaluated the performance of the following robots: A1 \cite{rudin2022learning}, Cassie \cite{rudin2022learning}, and MIT Humanoid \cite{jeon2023benchmarking} where each SNN is trained within $1,500,000$ iterations until convergence. 

\begin{figure}[tbp]
\setlength{\abovecaptionskip}{0.05cm}
\setlength{\belowcaptionskip}{-0.35cm}
\centerline{\includegraphics[width=0.985\columnwidth]{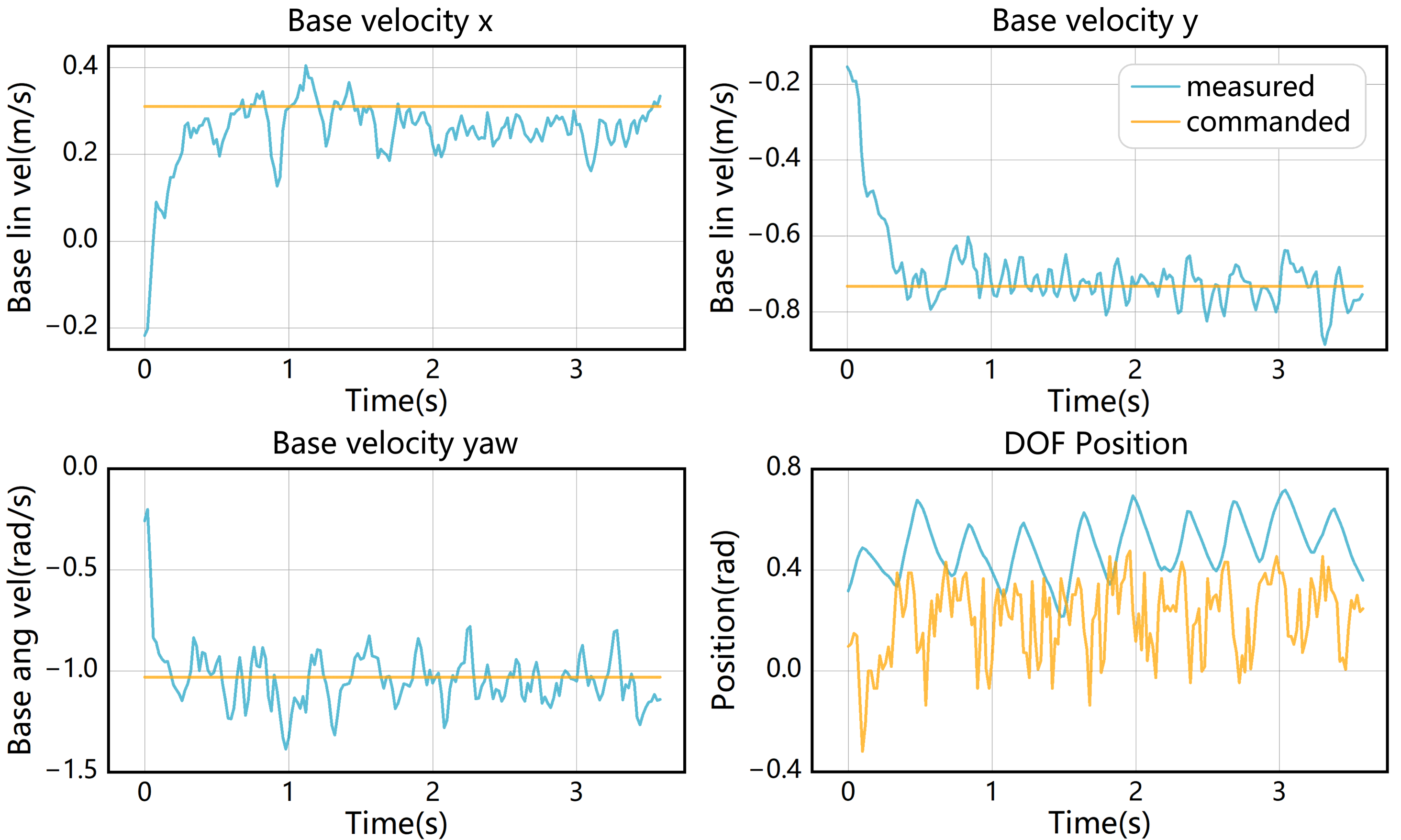}}
\caption{Four graphs illustrate the exceptional performance of the robot in command-following task.}
\label{fig5}
\end{figure}

\subsection{Simulation Setup}


In order to establish a varied array of environments, we incorporated multiple URDF files from recent studies. The files contain various models, including A1, Anymal-b, Anymal-c, Cassie, MIT Humanoid, and others. Once imported, we utilize the built-in fractal terrain generator of the Isaac Gym simulator to generate various environments for each of these models to introduce diversity. 
The policy functions at a control frequency of 500 Hz due to our SNN-based method, enabling quick and accurate system adjustments.

\subsection{Performances of High Frequency Control using SNNs}

We tested the aforementioned robots specifically on linear and angular velocity tracking tasks, using tracking velocity (linear and angular) as the primary reward. Penalties were applied for low base height, excessive acceleration, and instances of the robot falling, etc.
For A1, we conducted training and testing in several terrain environments, including pyramid stairs like terraces (upstairs and downstairs), pyramids with sloping surfaces, hilly terrains, terrains with discrete obstacles, and terrains covered with stepping stones. On the other hand, Cassie is solely trained in a trapezoidal pyramid environment and MIT Humanoid in a plain terrain.

\subsubsection{A1}

To explore the benefits of SNNs in high-frequency control, we increased the simulation environment's time step (dt) to 2.5 times that of the default ANNs task, achieving 500 Hz. ANNs are energy-constrained and typically reach only 100 Hz, lagging behind motor execution frequency. In contrast, SNNs may improve policy inference quality and enable real-time control due to their energy efficiency. If SNNs can match or exceed ANNs in high-frequency control, it would demonstrate their superiority in real-world environments. 
Figure \ref{fig5} shows the A1 robot effectively tracking velocity x and following the desired trajectory in complex terrain, with all velocities varying within $17\%$ of the designated value.


\subsubsection{Cassie}

\begin{figure}[tbp]
\setlength{\abovecaptionskip}{0.03cm}
\setlength{\belowcaptionskip}{-0.45cm}
\centerline{\includegraphics[scale=0.0575]{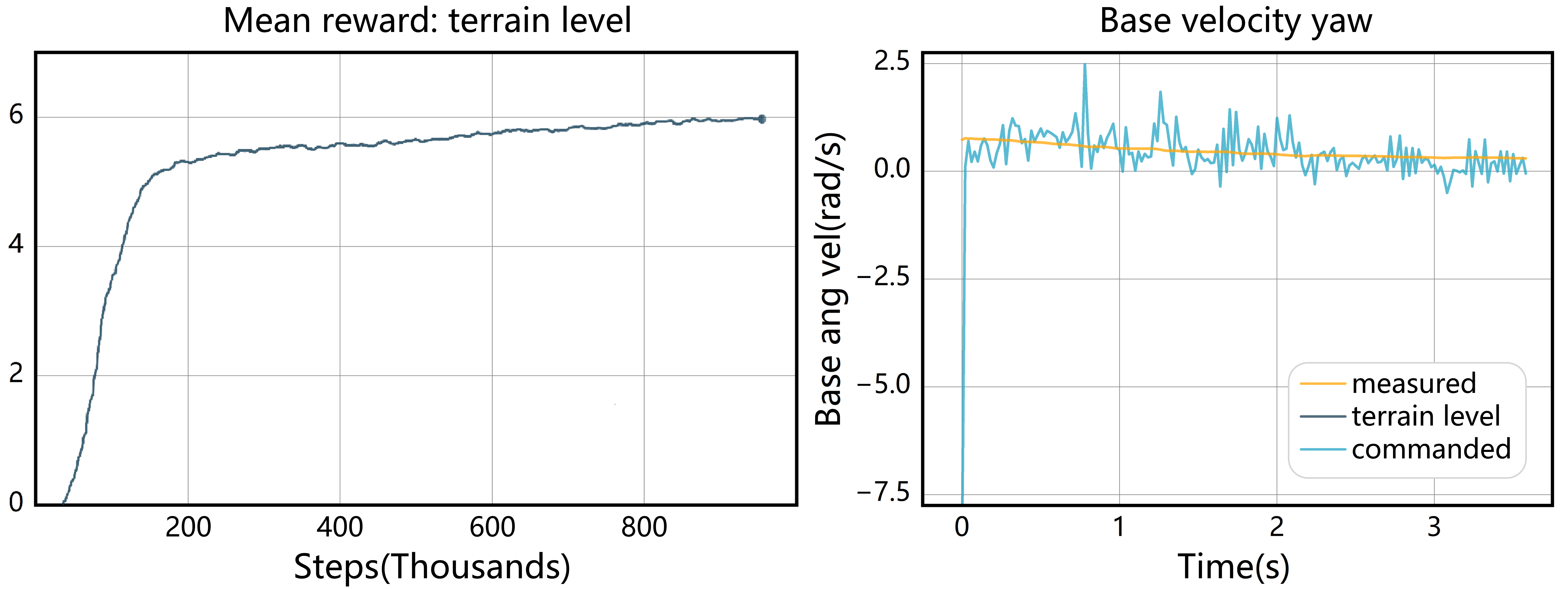}}
\caption{The first image showcases Cassie's remarkable capability to conquer complex terrain, as indicated by the terrain level value nearing 6. Additionally, the second figure demonstrates Cassie's impeccable tracking of the yaw axis's angular velocity, highlighting its stability while traversing complex terrain.}
\label{fig6}
\end{figure}

\begin{figure}[tbp]
\setlength{\abovecaptionskip}{0.05cm}
\setlength{\belowcaptionskip}{-0.2cm}
\centerline{\includegraphics[width=0.975\columnwidth]{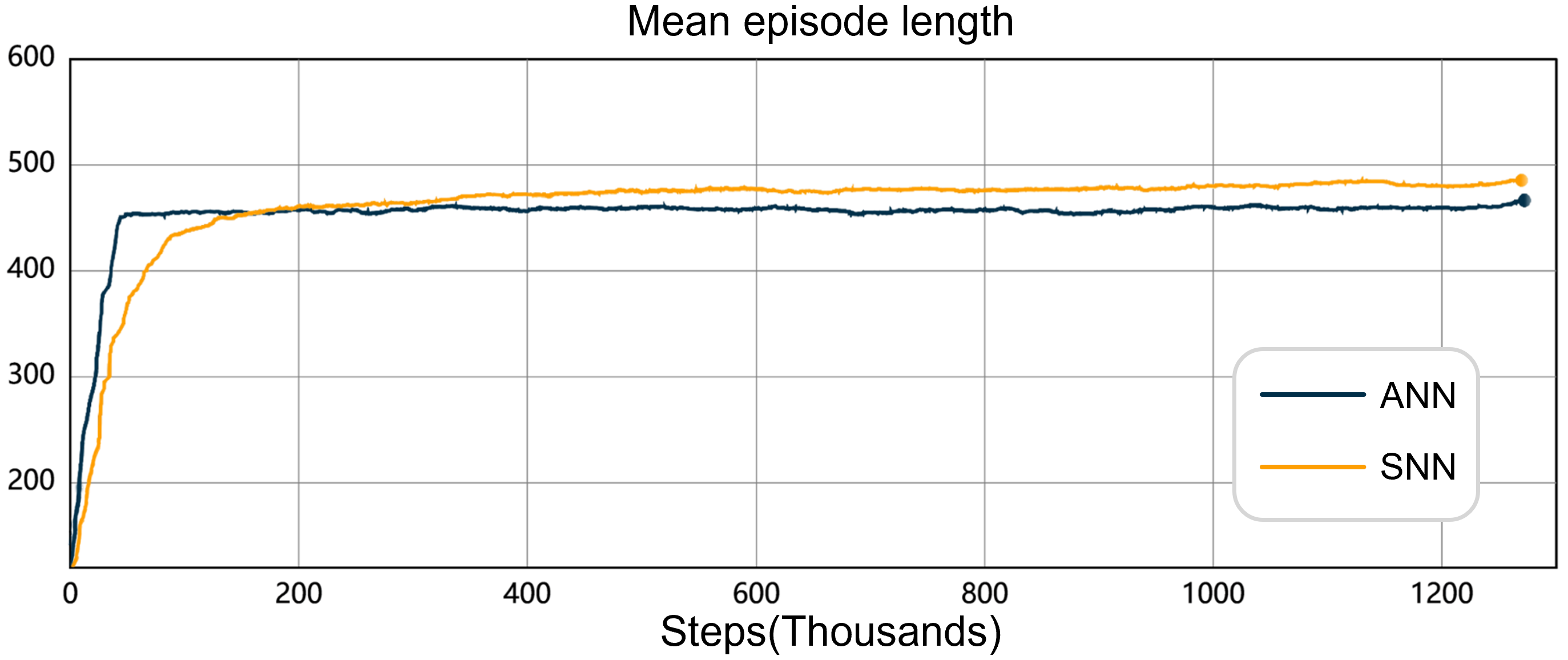}}
\caption{In experiments conducted on the MIT Humanoid, the SNN achieves a comparable level of approximation with ANN in multiple evaluation metrics and even surpasses ANN. Despite the fact that SNN takes longer to train, the SNN outperforms ANN in terms of mean episode length after training convergence, providing strong evidence for the exceptional robustness of our method in whole body control.
}
\label{fig7}
\end{figure}

In Cassie's experiments, the terrain level indicates the absolute value of the robot's elevation displacement (a value of 6 signifies reaching the top of the sixth-order pyramid). Figure \ref{fig6} highlights the stability of angular velocity, essential for maintaining effective control and balance while navigating diverse terrains. The robot's capability to reach the highest terrain levels further demonstrates its adaptability in conquering rugged landscapes.

\subsubsection{MIT Humanoid}

The MIT Humanoid training showcased the effectiveness of our spike-based approach. While it took longer to train than traditional ANNs, the results were impressive. In fact, they even surpassed the ANN in certain individual metrics, as clearly depicted in Figure \ref{fig7}. These findings strongly suggest that the SNN possesses advantages when it comes to control robustness. 
The performance demonstrated by our approach, whether through A1 and Cassie's agile traversal of challenging terrains or the MIT Humanoid's unrestricted running, is undeniably superior. 

\subsection{Ablation Experiment}

\begin{figure}[tbp]
\setlength{\abovecaptionskip}{0.05cm}
\setlength{\belowcaptionskip}{-0.43cm}
\centerline{\includegraphics[width=1.0\columnwidth]{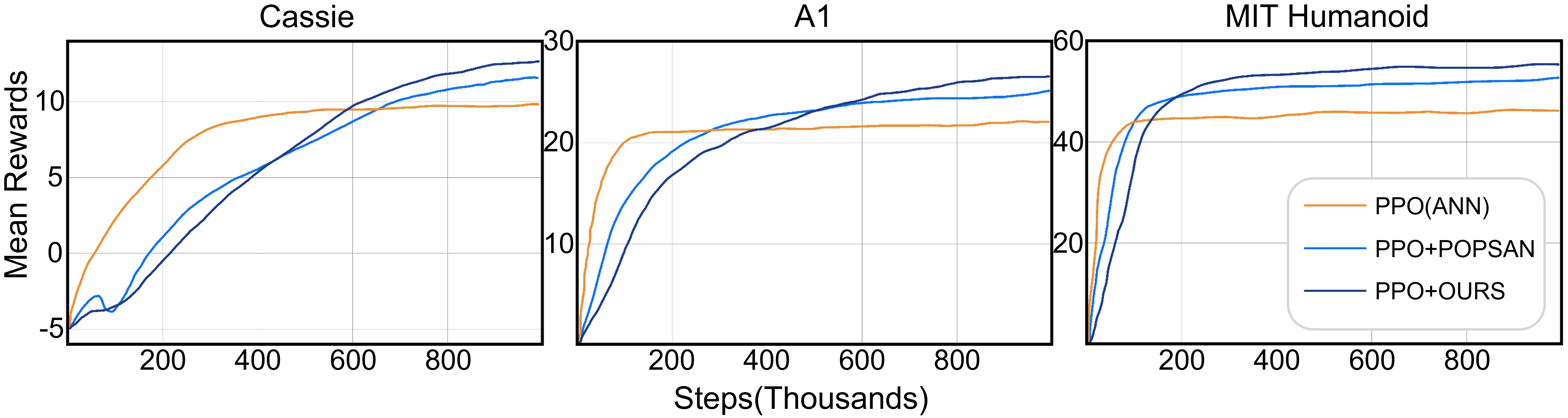}}
\caption{Results indicate that SNN-based DRL achieved performance comparable to ANN in high-frequency control scenarios, and our method yields the highest rewards. Plots are smoothed for clarity.
}
\label{fig8}
\end{figure}

We chose the Proximal Policy Optimization (PPO) algorithm \cite{schulman2017proximal} to evaluate our approach on three different robotic structures, assessing its universality. Figure \ref{fig8} shows that while the SNN-based DRL algorithm converges more slowly than the ANN-based DRL algorithm, it ultimately achieves comparable training results.


\subsection{Continuity Comparison}

SNNs exhibit greater robustness than ANNs because their thresholding mechanism filters noise, and their stochastic dynamics improve resilience to external disturbances.
We added Gaussian noise to the robot's movement commands, and tested the performance of the networks with two strategies, ANN and SNN, for sigma of 0.1, 0.2, and 0.3, respectively, using the following of the robot's y-axis linear velocity as an index. 
The experiment (Figure \ref{fig8}) shows that as noise increases, SNN demonstrates greater resilience, with measurements aligning closely to desired values and exhibiting fewer fluctuations. In contrast, ANN performs poorly; at a sigma of 0.3, it fails to support normal robot movement, leading to falls. The robot using the SNN, however, walks smoothly under the same noise conditions.

\begin{figure}[tbp]
\setlength{\abovecaptionskip}{0.05cm}
\setlength{\belowcaptionskip}{-0.1cm}
\centerline{\includegraphics[width=1.02\columnwidth]{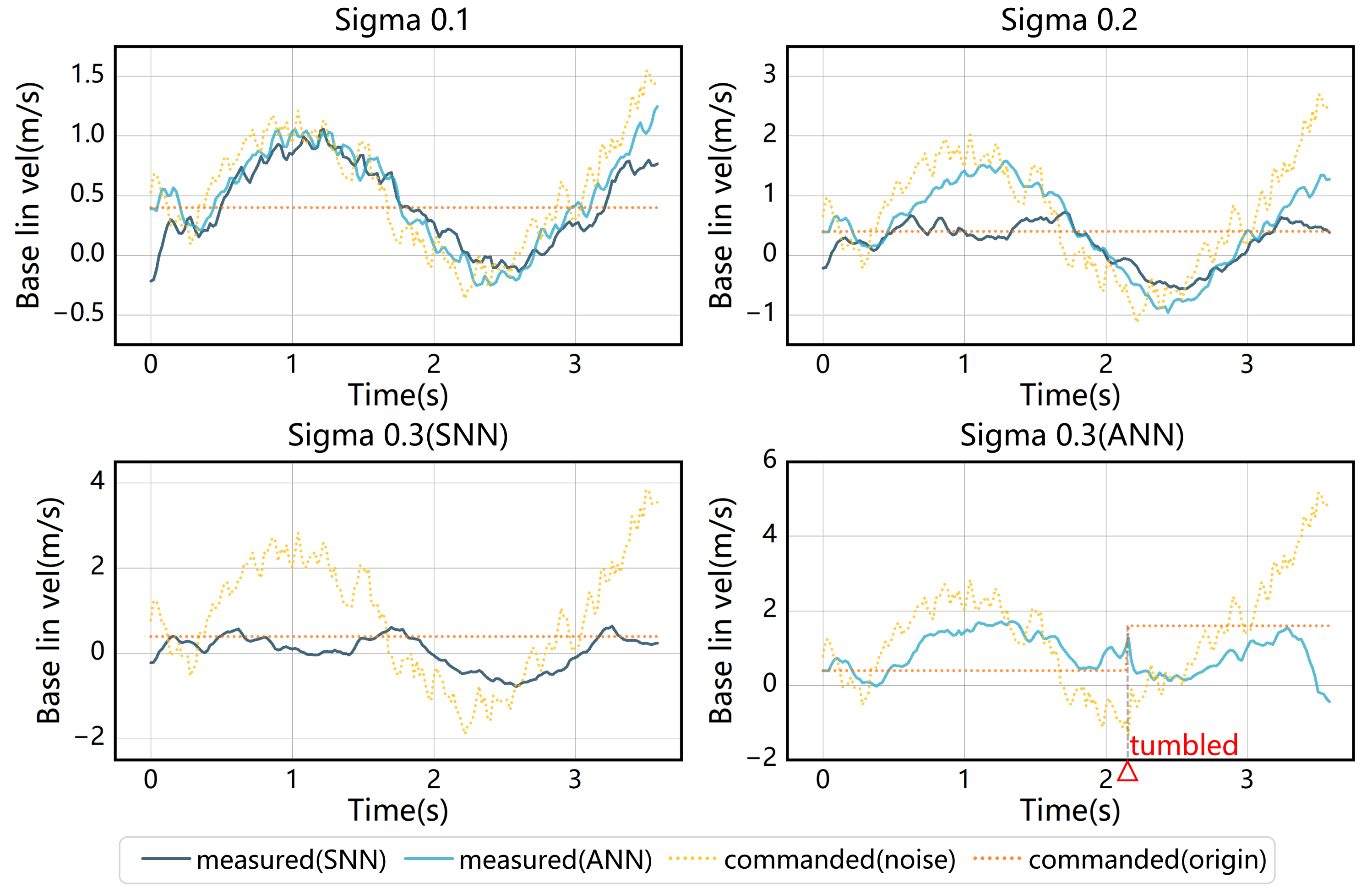}}
\caption{Linear velocity following case. The original command was set to 0.4, and Gaussian noise with a scaling factor of 0.1 was introduced. SNN consistently surpasses that of ANN when subjected to equivalent levels of noise, with this difference widening as the noise level intensifies. At a sigma of 0.3 for Gaussian noise, the robot with SNN as its strategy maintained stable command following and smooth walking, whereas the ANN version of the robot experienced instability and tumbled.
}
\label{fig9}
\end{figure}

\subsection{Estimation of Energy Consumption}

A key advantage of our SNN policy is its minimal energy usage. The assessment of energy consumption in a SNN is complex because the floating-point operations (FLOPs) in the initial encoder layer are MAC, whereas all other Conv or FC layers are AC. Building upon prior research\cite{hu2021advancing, kundu2021hire, yin2021accurate, yao2023attention} conducted by SNN, it is assumed that the data utilized for operations is represented in 32-bit floating-point format in 45nm technology\cite{yin2021accurate}, with $E_{MAC} =$ 4.6pJ and $E_{AC} =$ 0.9pJ. The energy consumption equations for SNN are shown below:
\begin{equation}
\setlength\abovedisplayskip{5pt}
\setlength\belowdisplayskip{5pt}
\begin{aligned}
E_{model} = & E_{MAC} \cdot FL^1_{SNNConv} + \\ & E_{AC} \cdot (\sum_{n=2}^NFL^n_{SNNConv} + \sum_{m=1}^MFL^m_{SNNFC})
\end{aligned}
\end{equation}
Experimental results (Table \ref{tab2}) indicate a significant energy efficiency improvement over conventional ANN architecture. Specifically, it achieves energy savings of $96.01\%$, $81.99\%$, and $58.86\%$ at $T_i$ = 1, 2 and 3, respectively. 
Each $T_i$ in the table represents the final stage, with $i$ set to $3$. The time dimension decreases by $1$ after each stage. The temporal shrinking between stages mainly involves simple fully connected layers, which are negligible compared to the final classifier. As the auxiliary classifier is not used during inference, overall energy consumption approximates that of the SNN main classifier.

\begin{table}[tbp]
\setlength\abovecaptionskip{0.08cm}
\renewcommand{\arraystretch}{1.3}
\caption{Energy Comparison($\times10^{-6}$ mJ)}
\vspace{-0.2cm}
\begin{center}
\begin{tabular}{cccc}
\hline
\specialrule{0em}{0.2pt}{0.1pt}
Method & Actor($T_i$=1) & 
Actor($T_i$=2) & Actor($T_i$=3) \\
\specialrule{0em}{0.1pt}{0.1pt}
\hline
\specialrule{0em}{1pt}{0.1pt}
ANN Model & 86.27 & 86.27 & 86.27 \\
SNN Model(\textbf{ours}) & \boldmath{$3.44$} & \boldmath{$15.54$} & \boldmath{$35.49$} \\
\specialrule{0em}{-1pt}{-1pt}
Energy Saving & 96.01\% & 81.99\% & 58.86\% \\
\hline
\end{tabular}
\label{tab2}
\end{center}
\vspace{-0.7cm}
\end{table}

\section{Conclusion}




This study presents the integration of a lightweight population coded SNN trained in a multi-stage method with history trajectory and imitation learning, which achieves performance comparable to ANNs, highlighting the versatility of SNNs in policy gradient-based DRL algorithms. This opens up new horizons for application of SNNs in various reinforcement learning tasks, including continuous, high dimensional control. Additionally, our approach offers significant advantages in energy efficiency, addressing signal noise, and high-frequency control. These advantages are significant for improving structural integrity and robustness in practical situations. It also has potential to reduce costs related to robot development and enables the implementation of advanced sensing systems in robotic platforms. 

By embracing SNNs, we unlock a realm of possibilities for future advancements in intelligent control systems, transcending traditional computational paradigms.



\bibliographystyle{IEEEtran}
\bibliography{ref}{}

\end{document}